\pdfoutput=1

\documentclass[11pt]{article}

\usepackage{coling}

\usepackage{times}
\usepackage{latexsym}

\usepackage[T1]{fontenc}

\usepackage[utf8]{inputenc}

\usepackage{microtype}

\usepackage{inconsolata}

\usepackage{graphicx}
\usepackage{inconsolata}
\usepackage{amsmath}
\usepackage{adjustbox}
\usepackage{multirow}
\usepackage{svg}
\usepackage[normalem]{ulem}
\useunder{\uline}{\ul}{}
\usepackage{makecell}
\usepackage{tablefootnote}
\usepackage{xcolor}
\definecolor{colorhigh}{HTML}{9900FF}
\definecolor{colorlow}{HTML}{660000}

\title{Multi-stage Training of Bilingual Islamic LLM for Neural Passage Retrieval}

\author{Vera Pavlova \\
  rttl labs, UAE \\
  \texttt{v@rttl.ai} \\}

\begin{document}
\maketitle
\begin{abstract}
This study examines the use of Natural Language Processing (NLP) technology within the Islamic domain, focusing on developing an Islamic neural retrieval model. By leveraging the robust XLM-R\textsubscript{Base} model, the research employs a language reduction technique to create a lightweight bilingual large language model (LLM). Our approach for domain adaptation addresses the unique challenges faced in the Islamic domain, where substantial in-domain corpora exist only in Arabic while limited in other languages, including English.

The work utilizes a multi-stage training process for retrieval models, incorporating large retrieval datasets, such as MS MARCO, and smaller, in-domain datasets to improve retrieval performance. Additionally, we have curated an in-domain retrieval dataset in English by employing data augmentation techniques and involving a reliable Islamic source. This approach enhances the domain-specific dataset for retrieval, leading to further performance gains.

The findings suggest that combining domain adaptation and a multi-stage training method for the bilingual Islamic neural retrieval model enables it to outperform monolingual models on downstream retrieval tasks.\footnote{A system is deployed at \url{https://rttl.ai/}}
\end{abstract}

\section{Introduction}
Despite the advancements in NLP technology, its application in the Islamic domain remains relatively limited. While various fields have harnessed NLP for tasks such as sentiment analysis, language translation, and chatbot development, the rich and complex textual resources within Islamic literature, such as the Holy Qur'an, Hadith, and scholarly articles, have not been fully leveraged.

Information retrieval (IR) plays a crucial role in the exploration of Islamic text. With the vastness of texts spanning centuries, efficient search methods are essential for scholars, researchers, and the general public. The ability to quickly locate specific passages, themes, or authors can significantly enhance understanding and facilitate deeper analysis. Moreover, given the intricate styles and diverse languages in which Islamic texts are written, advanced search tools can help bridge the gap between traditional scholarship and contemporary research needs. Effective retrieval not only saves time but also fosters a richer engagement with the cultural and intellectual heritage contained within Islamic literature \cite{bashir2023arabic}.
One of the significant challenges is the diversity of languages used in Islamic texts, including Arabic, English, Urdu, etc, which complicates the creation of robust NLP tools. Researching and developing multilingual retrieval tools could assist in accessing Islamic literature for Arabic and non-Arabic speakers.

\begin{figure}[t]
  \includegraphics[width=\columnwidth]{{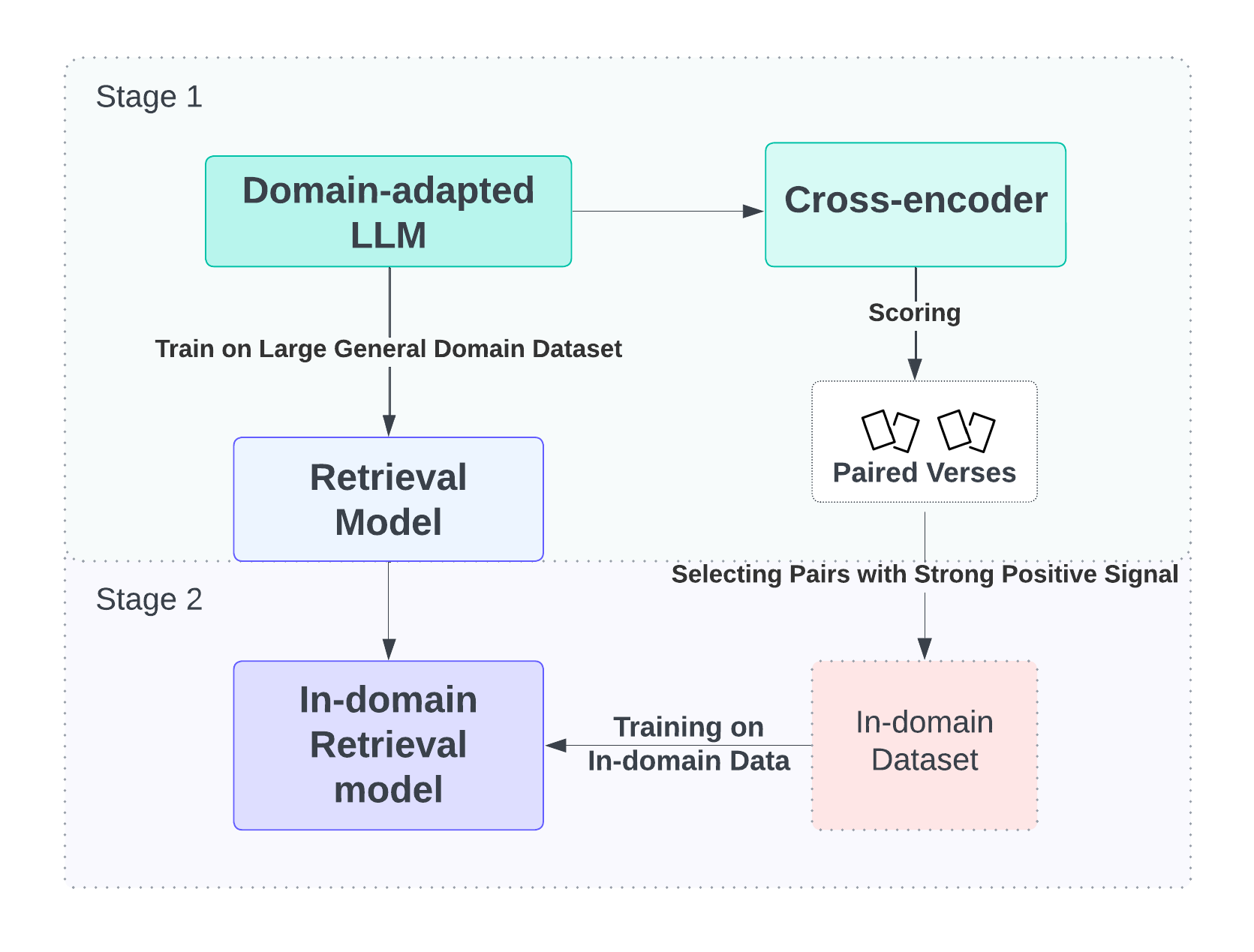}}
  \caption{Multi-stage training of Islamic neural retrieval model.}
  \label{fig:Figure 1}
\end{figure}

In this work, we study efficient ways to prepare a bilingual Islamic retrieval model. Addressing retrieval in both Arabic and English within the context of Islamic literature offers several significant benefits: firstly, it can increase the accessibility of Islamic literature to a broader audience. 
Classical Arabic (CA) is the language of the Holy Qur'an and plays a crucial role in conducting retrieval tasks involving sacred scripture. English is also widely used for search across various domains, including the Islamic field. 

Secondly, the use of multilingual or bilingual models enables cross-lingual transfer, which is crucial when there is insufficient data in some languages. English is a high-resource language, with a wealth of available corpora and pre-trained language models across various domains \citep{conneau-etal-2020-unsupervised}. On the other hand, Arabic has more advantageous resources within the Islamic domain than English due to the availability of large in-domain corpora such as OpenITI \citep{romanov2019openiti}. 
To benefit from both languages, we utilize the robust  XLM-R\textsubscript{Base} model, which has undergone extensive training on general domain corpora predominately in English and offers state-of-the-art performance on downstream tasks. 

We employ a language reduction technique \citet{abdaoui-etal-2020-load} that enables the creation of a robust, lightweight bilingual model, preserving most of the performance of the XLM-R\textsubscript{Base}. This model serves as the backbone for our retrieval system.  
It is known that retrieval models are sensitive to domain shifts, which can lead to a decline in performance \citep{thakur-2021-BEIR}. To address this issue, we perform domain adaptation using available text from Islamic literature and the OpenITI corpus in Arabic.

As a next step, we prepare a retrieval model using a dense retrieval approach \citep{karpukhin-etal-2020-dense, Izacard-unsupervised-dense}. Training a robust retrieval model requires a substantial amount of in-domain labeled data, which is currently not available in the Islamic domain. However, there are large general domain datasets available for training retrieval models. We propose a multi-stage training process for an Islamic neural retrieval model that leverages both the large general domain datasets as well as the small in-domain datasets. (see Figure ~\ref{fig:Figure 1}).

Additionally, we enhance our in-domain retrieval dataset in English by employing data augmentation techniques, which further improve the performance of the neural retrieval model. Our experiment showed that this approach improves the result on the evaluation dataset and outperforms strong monolingual baselines.

\section{Related Work}
Recent studies demonstrate that adapting existing LLMs pre-trained on general corpora for a new domain significantly improves performance on downstream tasks \citep{Lee-2019, Huang-ClinicalBERT}. The authors of the SciBERT model \cite{beltagy-etal-2019-scibert} showed that constructing a new Scivocab when pre-training SciBERT further enhances the performance of LLM. While pre-training a domain-specific model from scratch \citep{Gu-PubMedBERT} allows for the inclusion of domain-specific vocabulary, this approach is costly and often impractical when the domain-specific corpora are limited in size.
To avoid the random initialization of weights for new tokens and to expedite the pre-training process \citet{Poerner-etal-2020-inexpensive, sachidananda-etal-2021-efficient, pavlova-makhlouf-2023-bioptimus} experiment with introducing new vocabulary and pre-training domain-specific models using existing checkpoints.

There are several approaches to reducing model size. Research by \citet{sun-etal-2019-patient}, \citet{Tang-distillation}, \citet{Sanh-distillation}, and \citet{Li-distillation} has demonstrated that distilling transformer-based language models \citep{Vaswani-transformers} results in significant size reduction while maintaining adequate performance. 
Another approach is model quantization, as explored in studies by \citet{Guo-quantized}, \citet{Jacob-quantization}, \citet{bondarenko-etal-2021-understanding}, and \citet{tian-etal-2023-samp}. While quantization can help address model size issues, it often compromises performance.

In contrast, language reduction \citep{abdaoui-etal-2020-load} does not lead to substantial performance loss. This method decreases the model size by preserving the encoder weights and only trimming the embedding matrix, eliminating languages that are unnecessary for the specific task at hand.

The survey \citet {Zhao-dense-survey} provides a detailed overview of dense retrieval, including various model architectures and training approaches.
Other studies focusing on dense retrieval include \citet{karpukhin-etal-2020-dense}, \citet{qu-etal-2021-rocketqa}, \citet{ren-etal-2021-rocketqav2}.
\citet{thakur-etal-2021-augmented}, \citet{wang-etal-2021-tsdae-using} and \citet{wang-etal-2022-gpl} proposed a data augmentation technique to train retrieval models when there is little data for in-domain training. This approach involves creating synthetic data points that can mimic real in-domain scenarios, enriching the existing dataset, and bridging the gap between limited data availability and the need for high-quality model performance.

\section{Bilingual Islamic MLLM}
\label{sec:Preliminaries}

The application of cross-lingual transfer capabilities of MLLMs helped to solve important NLP tasks in low-resource languages \citep{devlin-etal-2019-bert, Lample-XLM}. \citet{conneau-etal-2020-unsupervised}  introduced the XLM-R and XLM-R\textsubscript{Base} with an increased model capacity trained on a large CommonCrawls corpus covering 100 languages.
He demonstrated that increasing model capacity and adding more languages improves cross-lingual performance on low-resource languages to a certain extent. However, beyond a certain point, the overall performance on both monolingual and cross-lingual benchmarks begin to decline, a notion that he referred to as the "curse of multilinguality."

\subsection{Size Reduction of LLM}
\label{sec: Size Reduction}

In this work, we want to explore the performance of the XLM-R\textsubscript{Base}  model after performing the language reduction technique, retaining only two languages (English and Arabic). We hypothesize that trimming the extended vocabulary of the XLM-R base model (250k) by removing languages not needed in the experiment will help reduce the model size, enhance model performance on downstream tasks, and facilitate domain adaptation. One of the main advantages of Language Reduction is that it reduces the number of languages by pruning only the embedding matrix while preserving all encoder weights. 
Unlike \citet{abdaoui-etal-2020-load}, our language reduction method consists of the following steps (see Figure ~\ref{fig: Figure 2}):
\begin{enumerate}
    \item  We select English and Arabic texts from a multilingual variant of the C4 corpus.
    \item Train a new SentencePiece BPE tokenizer using this corpus.
    \item We identify the intersection between the new tokenizer and the XLM-R\textsubscript{Base} tokenizer. \footnote{\url{https://huggingface.co/FacebookAI/xlm-roberta-base}} The tokens in this intersection, along with their corresponding weights, are copied to the new embedding matrix of the XLM-R2 model.
    \item The encoder weights from XLM-R\textsubscript{Base} are transferred directly to the new XLM-R2 model.
\end{enumerate}

\subsection{Domain Adaptation of MLLM}
\label{sec: Domain Adaptation}

Though language reduction allows us to benefit from the extensive training the XLM-R\textsubscript{Base}  model underwent, it gives us an LLM pre-trained for the general domain (XLM-R2). It is essential to note that the performance of retrieval models often declines when faced with domain shifts \citep{thakur-2021-BEIR}; since our focus in this study is on retrieval tasks and we apply this model as a backbone for retrieval, implementing domain adaptation is a crucial step to mitigate performance deterioration. 
In most domains, corpora to pre-train MLLMs are English-centric. However, we encounter a unique situation in the Islamic domain where significant domain-specific corpora are available in Arabic rather than English. Performing domain adaptation on a bilingual model brings certain advantages. On the one hand, the XLM-R\textsubscript{Base}  model was trained on extensive general domain English data, which helps to improve performance on Arabic tasks. On the other hand, the availability of larger Islamic corpora in Arabic also enables better results for domain-specific tasks in English.
The Open Islamicate Texts Initiative (OpenITI) \citep{romanov2019openiti} has provided a substantial corpus of 1 billion words for pre-training LLMs in Classical Arabic, the language of Arabic Islamic literature \citep{inoue-etal-2021-interplay, Malhas-CL-AraBERT}. While the available text in English is primarily composed of Tafseer and Hadith texts. Utilizing the OpenITI corpus can assist in creating a larger in-domain corpus for pre-training. To ensure the corpus is not overly biased towards Arabic, we randomly selected a subset of the OpenITI corpus containing approximately 50 million words. This subset was combined with the text of Hadeeth and Tafseer in English and Arabic, resulting in a total corpus size of 100 million words for domain adaptation.
The corpus size is relatively small; nevertheless, since the weights of the XLM-R2 model are initialized from the XLM-R\textsubscript{Base} model, we can employ continued pre-training. To tackle the word distribution shift, we incorporate new domain-specific vocabulary.

\begin{figure*}[t]
\centering
  \includegraphics[width=\textwidth]{{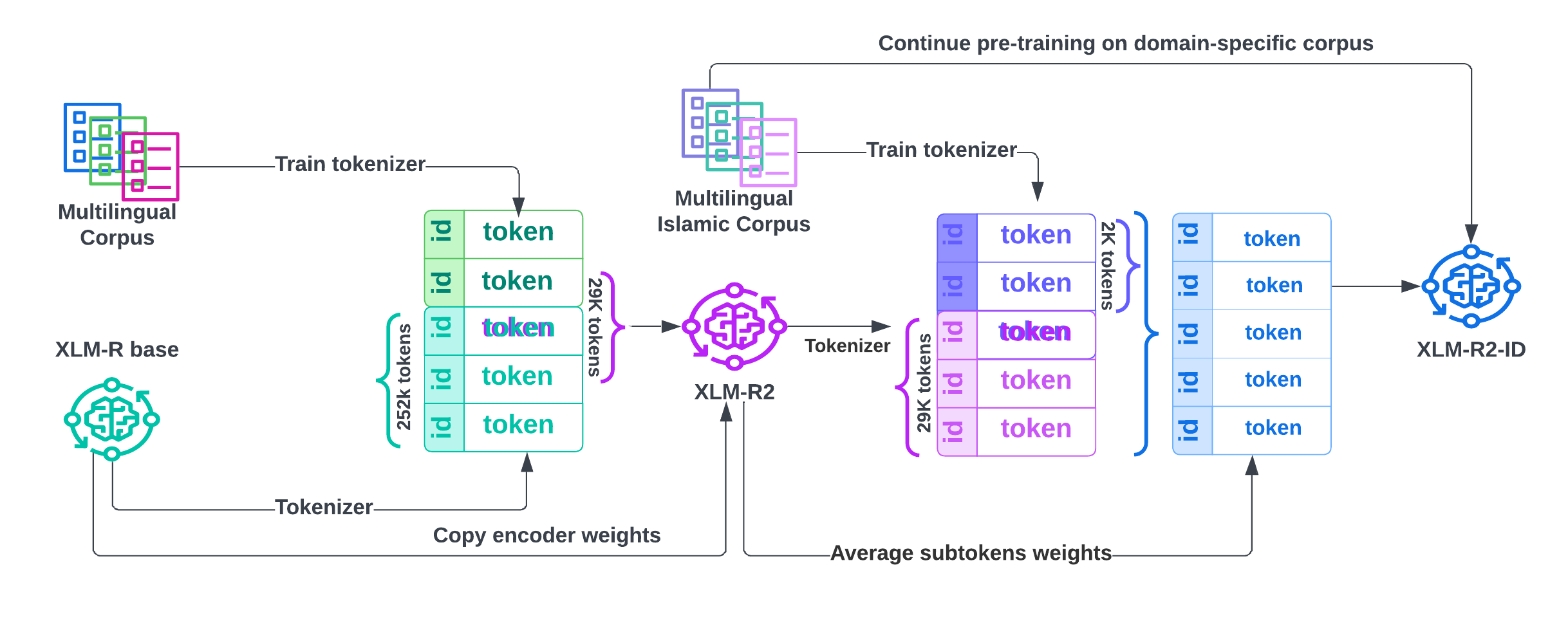}}
  \caption{Language Reduction (on the left) gives the XLM-R2 model that goes through Domain Adaptation (on the right) and brings the XLM-R2-ID model. Diagram from \citet{pavlova-makhlouf-2024-building} with permission.}
  \label{fig: Figure 2}
\end{figure*}

The steps of domain adaptation are the following (see Figure ~\ref{fig: Figure 2}):
\begin{enumerate}
    \item We train a new SentencePiece BPE tokenizer using a multilingual Islamic Corpus and identify the intersection between the new Islamic tokenizer and the XLM-R2 tokenizer. All the tokens outside of the intersection (2k tokens) are added to the embedding matrix of the XLMR-2 model (domain-specific vocabulary).
    \item The weights for new Islamic tokens are assigned by averaging existing weights of subtokens from the XLM-R2 model.
    \item We continue pre-training XLM-R2 using the domain-specific corpus, resulting in the XLM-R2-ID  (Islamic domain) model. For more details on the hyperparameters, refer to Appendix~\ref{sec:appendix}.
\end{enumerate}

\section{Domain-specific IR}
\subsection{Datasets, Metrics, and Training Approach}
\label{sec: Datasets}

For retrieval, we use a dense retrieval approach \citep{karpukhin-etal-2020-dense} using the sentence transformer framework that adds a pooling layer on top of LLM embeddings and produces fixed-sized sentence embedding \citep{reimers-gurevych-2019-sentence}. 

We utilize a sizeable general domain dataset (MS MARCO) for the first training stage in our multi-stage approach. The MS MARCO dataset consists of over half a million queries and is paired with a collection of 8.8 million passages \citep{Bajaj-MSMARCO}. The language dataset is English; \citet{Bonifacio-mmarco} released 13 machine-translated variants for 13 languages, including Arabic.
The transfer language for XLM-R\textsubscript{Base} is English, while -XLM-R2-ID has been adapted for the Islamic domain, primarily using Arabic. We will experiment with both English and Arabic as transfer languages to assess their effectiveness in addressing the IR task at hand.

The loss function is designed within the framework of contrastive learning, which helps create an embedding space that brings related queries and their relevant passages closer together while distancing queries and irrelevant passages \citep{Oord-Contrastive} , and formally defined as:

\begin{align*}
J_{\mathrm{CL}}&(\theta) = \\\frac{1}{M}\sum_{i=1}^M&\log\frac{\exp{\sigma(f_{\theta}(x^{(i)}), f_{\theta}(y^{(i)}))}}{\sum_{j=1}^M\exp{\sigma(f_{\theta}(x^{(i)}), f_{\theta}(y^{(j)}))}}
\end{align*}

where $\sigma$ is a similarity function (a cosine similarity), $f_{\theta}$ is the sentence encoder. 
To enhance training efficiency, we utilize in-batch negatives \citep{henderson2017efficient, gillick-etal-2019-learning, karpukhin-etal-2020-dense} (for hyperparameter details, see Appendix~\ref{sec:appendix}).

\textbf{In-domain training of retrieval model}.
Training a retrieval model on large-size general domain data would produce a robust model that can distinguish similar passages from dissimilar ones. However, training on a small amount of in-domain data can further enhance the performance \citep{wang-etal-2022-gpl, lu-etal-2021-multi}.
For in-domain training for the second stage, we use QUQA \citep{alnefaie-etal-2023-haqa}, an Arabic question-answering (QA) dataset based on the Holy Qur'an. It has 3382 pairs, including 1166 pairs from AyaTEC \citep{malhas2020ayatec}, which we exclude as we use them for evaluation (see below). Some questions relate to more than one verse. We take each relation as a separate anchor-positive pair, which gives us 3252 pairs. The dataset has only an Arabic version; we curate English in-domain training data to improve the in-domain training.  

The challenge of limited domain-specific data can often be addressed by augmenting the training data using various methods. These methods include generating synthetic data \citep{tanaka2019data}, paraphrasing with synonyms \citep{wei-zou-2019-eda}, sampling and recombining new training pairs \citep{thakur-etal-2021-augmented}, employing round-trip translation \citep{yu2018qanet, xie2020unsupervised}, and utilizing denoising autoencoders \citep{wang-etal-2021-tsdae-using}. However, these techniques can distort the data, which is not ideal for religious and heritage datasets.
To prevent data distortion, we create anchor-positive pairs based on the verse relations mentioned in Tafseer Ibn Katheer. This method facilitates the creation of relevant, verified, high-quality in-domain data without costly human annotations.

\begin{itemize}
\item First, we pair all the verses that have relation mentioned in Tafseer Ibn Katheer.
\item Next, we filter out the pairs that may not be interpreted by the model as indicating a strong positive correlation. To filter out these pairs, we scored them using a cross-encoder model trained using the XLM-R2-ID model (for details on cross-encoder training see Appendix~\ref{sec:appendix}). Cross-encoder is a powerful but expensive approach to identifying similar pairs. Though they are suboptimal to apply in solving real-world retrieval tasks due to high computational overhead, they can help in data augmentation, distillation, and re-ranking without enduring considerable domain shift \citep{humeau2020polyencoders, wang-etal-2022-gpl}. Cross-encoder helps to filter out pairs with low similarity scores, leaving us with 2133 pairs for in-domain training.
\item Lastly, we combine Arabic QUQA pairs with English pairs, which results in 5385 pairs for training. 
\end{itemize}

Unlike \citet{pavlova-2023-leveraging}, we do not create hard negatives for training the retrieval model. Instead, we apply the same contrastive learning framework, utilizing in-batch negatives as described earlier, to train on in-domain data. We ensure that there are no duplicate entries within the same batch and sample from English and Arabic in-domain datasets in proportion to their respective sizes. This approach allows all samples from each dataset to be utilized.

For evaluation, we combined the train and development split of the QRCD (Qur'anic Reading Comprehension Dataset) \citep{malhas2020ayatec} and converted it to the IR dataset (169 queries in total for testing). The QRCD dataset is in Arabic; to conduct evaluations in English, we utilized verified translations of this dataset into English.
We use the Holy Qur'an text (Arabic) and Sahih International translation (English) as retrieval collections.\footnote{\url{https://tanzil.net/trans/}}
The QRCD is designed to retrieve passages composed of verses from the Holy Qur'an. The Holy Qur'an texts mentioned above are organized according to the passages based on the QRCD.
We evaluate the models' performance using decision support metric Recall@100 and the order-aware metric MRR@10 (MS MARCO's official metric). 

\subsection{Baselines and Models}
We use two monolingual models as our baselines. For English, we use 
ST/all-mpnet-base-v2, a robust monolingual model trained with contrastive learning objectives on 1B sentence pairs. \footnote{\url{https://huggingface.co/sentence-transformers/all-mpnet-base-v2}} For Arabic, we use CL-AraBERT \citep{Malhas-CL-AraBERT}, which was pre-trained on OpenITI corpus, and we fine-tuned as a retrieval model on Arabic MS MARCO and in-domain Arabic data using the same training loss described above (for hyperparameters see Appendix~\ref{sec:appendix}). 
This choice of baselines serves two purposes. First, it enables us to evaluate how our bilingual model performs compared to monolingual models. Second, comparing our model against a strong retrieval model that is not domain-adapted (ST/all-mpnet-base-v2) allows us to assess the effects of domain adaptation. Additionally, contrasting it with a retrieval model trained using CL-AraBERT—adapted with the full OpenITI dataset, which consists of about 1 billion words — will help us evaluate the potential of utilizing a smaller corpus for domain adaptation.

As previously mentioned in Section ~\ref{sec: Datasets}, we experimented with MS MARCO in both English and Arabic to select which language performs better for retrieval tasks in the Islamic domain. We trained two models, XLM-R2-ID-EN and XLM-R2-ID-AR, which correspond to the first stage of our multi-stage training approach. We selected the model that exhibited superior performance on the evaluation dataset for use in the second stage.

In the second stage, where we train on in-domain data, we developed the  XLM-R2-ID-AR-in-domain. To evaluate the impact of multi-stage training, we also produced the XLM-R2-ID-in-domain model, which was trained solely on in-domain data without using MS MARCO.
We aimed to analyze the performance of the XLM-R2-ID model after implementing domain adaptation and language reduction. To facilitate this analysis, we prepared four additional models trained from the XLM-R\textsubscript{Base} model for comparison with four models trained from the XLM-R2-ID. These include two models trained on the MS MARCO dataset in English and Arabic (XLM-R-EN and XLM-R-AR), a model trained in a multi-stage approach (XLM-R-AR-in-domain), and a model trained solely on in-domain data (XLM-R-in-domain).

\begin{table}[t]
\begin{adjustbox}{width=\columnwidth,center}
\begin{tabular}{lcc}
\hline
\textbf{Models}                       & \textbf{EN}    & \textbf{AR}
\\ \hline
\textbf{ST/all-mpnet-base-v2}         &{\ul0.388}      & -
\\
\textbf{CL-AraBERT-indomain}          & -              & {\ul0.512}
\\ \hline
\textbf{XLM-R-in-domain}             &  0.211         &  0.218
\\
\textbf{XLM-R-EN}                    & 0.302          & 0.267
\\
\textbf{XLM-R-AR}                    & 0.287          & 0.264
\\
\textbf{XLM-R-AR-in-domain}          & 0.308          & 0.316
\\ \hline
\textbf{XLM-R2-ID-in-domain}         & 0.348           & 0.465      
\\
\textbf{XLM-R2-ID-EN}                & 0.329          & 0.416     
\\
\textbf{XLM-R2-ID-AR}               & 0.387           & 0.498      
\\
\textbf{XLM-R2-ID-AR-in-domain}      &\textbf{0.441}  &\textbf{0.534}
\\ \hline
\end{tabular}
\end{adjustbox}
\caption{Model performance for MRR@10.}
\label{tab:table 1}
\end{table}

\begin{table}[t]
\begin{adjustbox}{width=\columnwidth,center}
\begin{tabular}{lcc}
\hline
\textbf{Models}                       & \textbf{EN}    & \textbf{AR}
\\ \hline
\textbf{ST/all-mpnet-base-v2}         &{\ul0.619}      & -
\\
\textbf{CL-AraBERT-indomain}          & -              & {\ul0.756}
\\ \hline
\textbf{XLM-R-in-domain}             &  0.451         &  0.462
\\
\textbf{XLM-R-EN}                    &  0.492         & 0.496
\\
\textbf{XLM-R-AR}                    & 0.493          & 0.541
\\
\textbf{XLM-R-AR-in-domain}          &  0.528         & 0.584
\\ \hline
\textbf{XLM-R2-ID-in-domain}         &  0.592          & 0.706    
\\
\textbf{XLM-R2-ID-EN}                &  0.571         &  0.675    
\\
\textbf{XLM-R2-ID-AR}                &  {\ul0.619}    & 0.72  
\\
\textbf{XLM-R2-ID-AR-in-domain}      &\textbf{0.646}  &\textbf{0.766}
\\ \hline
\end{tabular}
\end{adjustbox}
\caption{Model performance for Recall@100.}
\label{tab:table 2}
\end{table}

\subsection{Model Comparison}
\label{sec: Comparison}
For our model comparison, we will examine four key aspects: First, we will compare model performance against a baseline. Second, we will assess how models trained from domain-adapted XLM-R2-ID models perform compared to those trained from general domain  XLM-R\textsubscript{Base}.  Third, we will evaluate model performance that was trained in a multi-stage approach against training conducted solely with in-domain datasets (only stage two) or only with large general domain datasets (only stage one). Finally, we will analyze how the models perform on the evaluation dataset in English versus the evaluation dataset in Arabic.

In Tables  ~\ref{tab:table 1} and  ~\ref{tab:table 2}, the best-performing model is in bold, and the second-best is underlined. 
In terms of MRR@10 for English (see Table ~\ref{tab:table 1}), we see that most models perform worse than the baseline monolingual model, all-mpnet-base-v2 (0.388), with the exception of XLM-R2-ID-AR-in-domain (0.441). This model, which was trained using a multi-stage approach, outperforms the baseline. A similar trend is observed for MRR@10 in Arabic.

Table ~\ref{tab:table 1} shows that the second stage of in-domain training provides significant benefits for English, resulting in a 12\% performance improvement (increasing from 0.387 to 0.441). For Arabic, the second stage yields an improvement of approximately 6\%. The same is true for Recall@100 (see Table  ~\ref{tab:table 2}); only the models that utilized a multi-stage training approach were able to surpass a strong monolingual baseline, but the improvement was less pronounced than that observed in MRR@10.

\begin{figure}[t]
  \includegraphics[width=\columnwidth]{{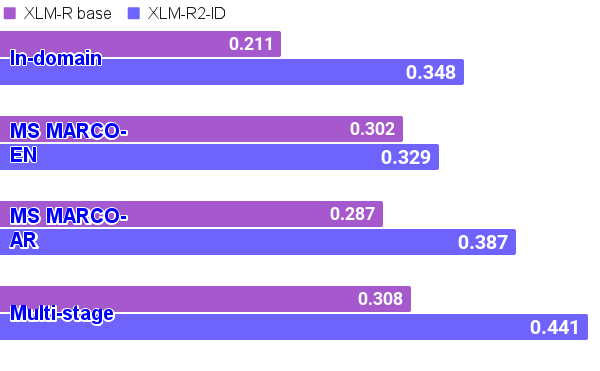}}
  \caption{Comparison of the performance of the retrieval models trained from XLM-R\textsubscript{Base} and XLM-R2-ID for MRR@10 in English.}
  \label{fig:figure 3}
    \includegraphics[width=\columnwidth]{{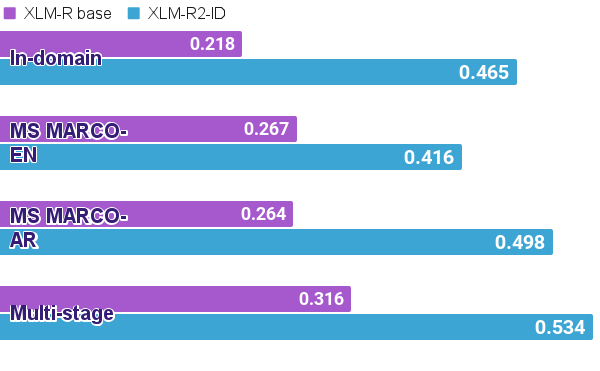}}
  \caption{Comparison of the performance of the retrieval models trained from XLM-R\textsubscript{Base} and XLM-R2-ID for MRR@10 in Arabic.}
  \label{fig:figure 4}
\end{figure}

\begin{figure}[t]
  \includegraphics[width=\columnwidth]{{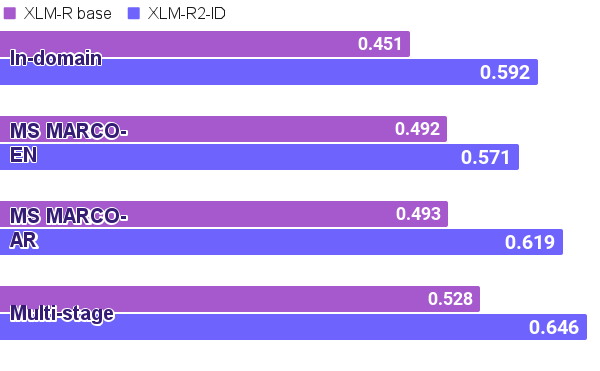}}
  \caption{Comparison of the performance of the retrieval models trained from XLM-R\textsubscript{Base} and XLM-R2-ID for Recall@100 in English.}
  \label{fig:figure 5}
    \includegraphics[width=\columnwidth]{{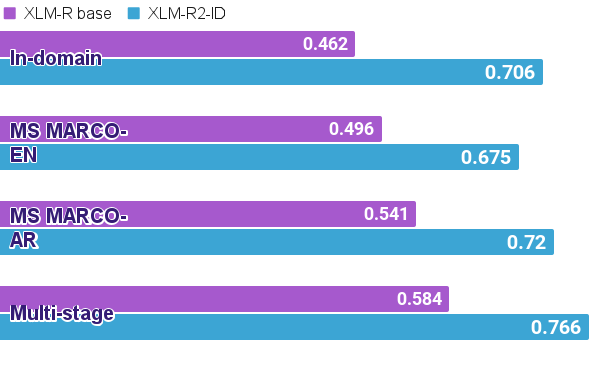}}
  \caption{Comparison of the performance of the retrieval models trained from XLM-R\textsubscript{Base} and XLM-R2-ID for Recall@100 in Arabic.}
  \label{fig:figure 6}
\end{figure}

\begin{figure}[t]
  \includegraphics[width=\columnwidth]{{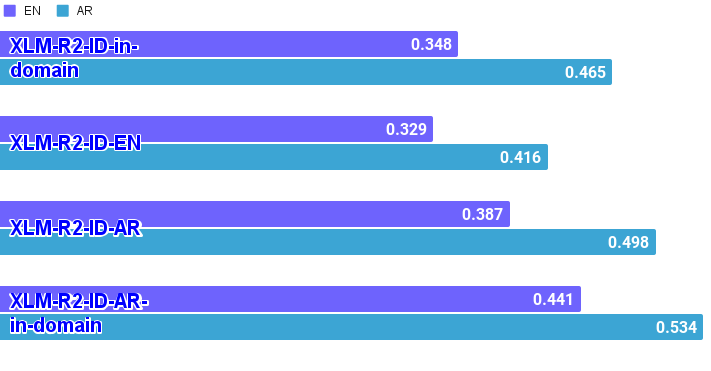}}
  \caption{Comparison of results between Arabic and English for MRR@10.}
  \label{fig:figure 7}
    \includegraphics[width=\columnwidth]{{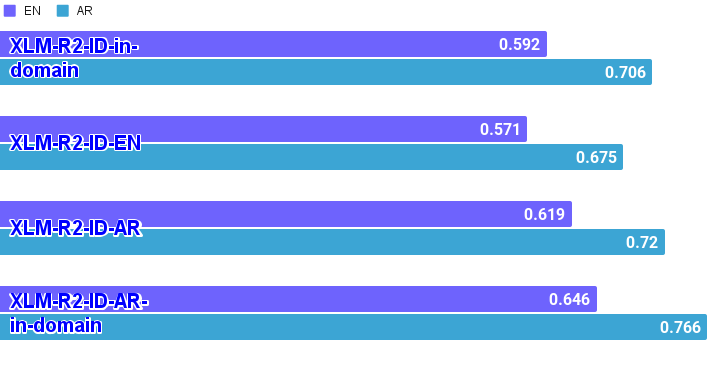}}
  \caption{Comparison of results between Arabic and English for Recall@100.}
  \label{fig:figure 8}
\end{figure}

Figures ~\ref{fig:figure 3} and ~\ref{fig:figure 4} illustrate the performance comparison between all stages of the models trained from  XLM-R\textsubscript{Base} and the XLM-R2-ID for MRR@10, and Figures ~\ref{fig:figure 5} and ~\ref{fig:figure 6} for Recall@100. It is evident that all XLM-R2-ID models outperform the  XLM-R\textsubscript{Base} models in both Arabic and English, with a particularly significant difference in performance observed in Arabic. This indicates that domain adaptation has been especially beneficial for the Arabic language. 

Another important observation is that  XLM-R\textsubscript{Base} models do not respond to multi-stage training as effectively as models based on XLM-R2-ID. Additionally, training solely on in-domain data results in competitive models; however, these still fall short compared to models trained using a multi-stage approach, similar to those trained only on general domain data.

Interestingly, although machine-translated, training on the Arabic version of MS MARCO yields better results than training on the original English version of MS MARCO. This trend holds true for evaluations in both Arabic and English. Despite starting the domain adaptation process from the  XLM-R\textsubscript{Base}  model, which is an English-centric model, training on a small amount of Arabic data and adding domain-specific vocabulary significantly improved the results for the Arabic language.

As a final part of our comparison, we closely examine the performance of models using English and Arabic datasets. Figure ~\ref{fig:figure 7} (MRR@10) and Figure ~\ref{fig:figure 8} (Recall@100) demonstrate that the results for Arabic are superior to those for English across all models. The disparity is particularly pronounced in MRR@10, while the difference is less significant for Recall@100.

Overall, our experiments and comparisons reveal several important findings: domain adaptation of LLMs, even with a small corpus, significantly contributes to improved performance on downstream tasks. However, adaptation alone was insufficient to surpass strong monolingual baselines. Instead, the multi-stage training approach enhanced the results, allowing us to outperform the baselines.

Additionally, the XLM-R2-ID-in-domain model outperformed the retrieval model trained on CL-AraBERT, which was developed on a larger corpus. This suggests that even a small corpus can be effective, especially when leveraging strong XLM-R\textsubscript{Base} weights for a warm start.

\begin{table}[t]
\begin{adjustbox}{width=\columnwidth,center}
\begin{tabular}{lcc}
\hline
\textbf{Models}                       & \textbf{EN}    & \textbf{AR}
\\ \hline
\textbf{XLM-R2-ID-AR-in-domain}       & \textbf{0.441}  & \textbf{0.534}   
\\
\textbf{XLM-R2-ID*-AR-in-domain}       & 0.414          & 0.498 
\\
\textbf{XLM-R2-ID-in-domain*}          & 0.381        & 0.521
\\ \hline
\end{tabular}
\end{adjustbox}
\caption{Model performance for MRR@10 with two ablated models.}
\label{tab:table 3}
\end{table}

\begin{table}[t]
\begin{adjustbox}{width=\columnwidth,center}
\begin{tabular}{lcc}
\hline
\textbf{Models}                       & \textbf{EN}    & \textbf{AR}
\\ \hline
\textbf{XLM-R2-ID-AR-in-domain}      &  \textbf{0.646}  & \textbf{0.766}   
\\
\textbf{XLM-R2-ID*-AR-in-domain}        &  0.643         &  0.739
\\
\textbf{XLM-R2-ID-in-domain*}            &  0.622         & 0.751
\\ \hline
\end{tabular}
\end{adjustbox}
\caption{Model performance for Recall@100 with two ablated models.}
\label{tab:table 4}
\end{table}

\section{Ablation Study}

In this section, we conduct an ablation study on various aspects of our highest-performing model, XLM-R2-ID-AR-in-domain, which was trained using domain adaptation with extended corpus and a multi-stage approach with augmented in-domain data. 
To better understand the relative importance of each component, we examine the effects of removing each one individually.

First, we removed the domain adaptation that involved extending the Islamic corpus with OpenITI. We then used only the available Islamic texts from Hadith and Tafseer in English and Arabic, totaling 50 million words. In Tables ~\ref{tab:table 3} and ~\ref{tab:table 4}, we present the performance of the model XLM-R2-ID*-AR-in-domain, which was trained without OpenITI but still employed a multi-stage approach. The results show that MRR@10 decreased by approximately 6\% for both Arabic and English. For Recall@100, the difference in performance for English was relatively small, whereas for Arabic, it was about 3.5\%.

Next, we removed the augmentation of the in-domain corpus with English data, resulting in the model XLM-R2-ID-in-domain*. As shown in Table ~\ref{tab:table 3}, this led to a more substantial decline in performance for English — around 14\% - while the decrease for Arabic was only about 2.43\%. Again, for Recall@100 (Table ~\ref{tab:table 4}), the difference was less pronounced. This indicates that augmenting the in-domain data with English has a significant impact, especially for performance on retrieval task in English. 

Moreover, our findings suggest that expanding the pre-training corpus with OpenITI improved results for both Arabic and English. Since adding additional texts amounting to 50 million words does not significantly prolong pre-training time, we recommend this approach; however, as we demonstrated in the section ~\ref{sec: Comparison}, even without pre-training on an extended corpus of 1 billion words, using a relatively small corpus for domain adaptation can still yield significant improvements.

\section {Conclusion}
This study emphasizes the importance of leveraging domain adaptation to enhance the performance of LLMs on downstream tasks such as retrieval. By combining language reduction with domain adaptation applied to the XLM-R\textsubscript{Base} model, we developed a lightweight bilingual Islamic LLM (XLM-R2-ID). This model underwent a multi-stage training process and demonstrated improved performance on retrieval tasks, surpassing monolingual models on the evaluation datasets.

Moreover, incorporating an augmented in-domain dataset in English further enhanced the performance of the retrieval model during the second training phase.

Overall, our research demonstrates that combining domain adaptation of LLMs with multi-stage training of neural retrieval models leads to improved results in downstream tasks such as IR.

\section*{Limitations}
One of the limitations of our study is that we conducted experiments only in English and Arabic; experiments that involve other languages may vary. Additionally, we used machine-translated datasets. While machine translation has not yet reached the quality of expert human translation, our use of the Arabic machine-translated version of MS MARCO has shown promising results.  

\section*{Ethics Statement}
We do not anticipate any considerable risks associated with our work. 
The data and other related resources in this work are publically available, and no private data is involved.

\section*{Acknowledgment}
This work would not have been possible without my colleague Mohammed Makhlouf. We extend our thanks to the anonymous Reviewers for their valuable feedback.

\bibliography{anthology,custom}

\appendix

\section{Appendix}
\label{sec:appendix}

\begin{table}[h]
\begin{adjustbox}{width=\columnwidth,center}
\begin{tabular}{ccc}
\hline
\textbf{Computing Infrastructure}    & \multicolumn{2}{c}{1x H100 (80 GB)}    \\ \hline
\multicolumn{2}{c}{\textbf{Hyperparameter}} & \textbf{Assignment}                     \\ \hline
\multicolumn{2}{c}{number of epochs}        & 60 \\ \hline
\multicolumn{2}{c}{batch size}              & 128                             \\ \hline
\multicolumn{2}{c}{maximum learning rate}   & 0.0005                        \\ \hline
\multicolumn{2}{c}{learning rate optimizer} & Adam                                    \\ \hline
\multicolumn{2}{c}{learning rate scheduler} & None or Warmup linear                   \\ \hline
\multicolumn{2}{c}{Weight decay}            & 0.01                                    \\ \hline
\multicolumn{2}{c}{Warmup proportion}       & 0.06                                    \\ \hline
\multicolumn{2}{c}{learning rate decay}     & linear                                  \\ \hline
\end{tabular}
\end{adjustbox}
\caption{Hyperparameters for pre-training of XLM-R2-ID model.}
\label{tab:appendix-table-a}
\end{table}

\begin{table}[!htbp]
\begin{adjustbox}{width=\columnwidth,center}
\begin{tabular}{ccc}
\hline
\textbf{Computing Infrastructure} & \multicolumn{2}{c}{1x H100 (80 GB)} \\ \hline

\multicolumn{2}{c}{\textbf{Hyperparameter}}     & \textbf{Assignment}           \\ \hline
\multicolumn{2}{c}{number of epochs}            & 1                            \\ \hline
\multicolumn{2}{c}{batch size}                  & 32                             \\ \hline
\multicolumn{2}{c}{learning rate}               & 2e-5                          \\ \hline
\multicolumn{2}{c}{pooling}                     & mean                         \\ \hline

\end{tabular}
\end{adjustbox}
\caption{Hyperparameters for training retrieval models.}
\label{tab:appendix-table-b}
\end{table}

\textbf{Cross-encoder training details}

In a cross-encoder architecture, a pair of sentences are simultaneously fed into a transformer-like model, allowing attention to be applied across all tokens to generate a similarity score. The model is trained using triples provided by MS MARCO, starting from the XLM-R2-ID model checkpoint, with a classification task and employing Cross Entropy Loss.

Although this approach does not enable end-to-end information retrieval and involves significant computational overhead, it often outperforms other methods in many information retrieval (IR) tasks. Additionally, it can be utilized for mining hard negatives, data augmentation, and reranking. 

\end{document}